\documentclass[letterpaper,10pt,journal]{IEEEtran}
\usepackage{amsmath,amsfonts}
\usepackage{algorithmic}
\usepackage{algorithm}
\usepackage{array}
\usepackage[caption=false,font=normalsize,labelfont=sf,textfont=sf]{subfig}
\usepackage{textcomp}
\usepackage{stfloats}
\usepackage{url}
\usepackage{verbatim}
\usepackage{graphicx}
\usepackage{siunitx}
\usepackage{amsthm}
\usepackage{hyperref}
\newtheorem{definition}{Definition}
\def\BibTeX{{\rm B\mathrm{K}ern-.05em{\sc i\mathrm{K}ern-.025em b}\mathrm{K}ern-.08em
    T\mathrm{K}ern-.1667em\lower.7ex\hbox{E}\mathrm{K}ern-.125emX}}
\usepackage{balance}

\newcommand{\K}{\mathrm{K}}
\newcommand{\M}{\mathrm{M}}
\newcommand{\B}{\mathrm{B}}
\newcommand{\KXY}{\mathrm{K_{xy}}}
\newcommand{\DXY}{\mathrm{\Delta_{xy}}}

\begin{document}
\title{Stability Boundaries and Motor Performance in Delayed Robot-Mediated Dyadic Interactions}
\author{Mingtian Du$^{1}$, Suhas Raghavendra Kulkarni$^{1}$, Simone Kager$^{2}$, Erkan Kayacan$^{1}$, Domenico Campolo$^{1}$
\thanks{*Funding information will be included upon acceptance. Links to supplementary materials, including datasets and appendices, will be provided in the final version.}
\thanks{$^{1}$M. Du, S. R. Kulkarni, E. Kayancan, and D. Campolo are with Robotics Research Centre, School of Mechanical and Aerospace Engineering, Nanyang Technological University, Singapore 639798
        {\tt\small (corresponding author (D. Campolo): d.campolo@ntu.edu.sg).}}
\thanks{$^{2}$S. Kager is with the Department of Computer Science and Artificial Intelligence, University of Technology Nuremberg, Germany 90461.}
}
\maketitle

\begin{abstract}
This paper establishes analytical stability boundaries for robot-mediated human–human (dyadic) interaction systems, subject to haptic communication under network-induced time delays. Bypassing conservative approximations, we employ a frequency-domain zero-crossing methodology to extract explicit stability limits based on the robotic hardware dynamics and coupling stiffness. To demonstrate the scalability of this mathematical framework, we extend the analysis from an elastic coupling to a highly complex, asymmetric virtual proxy topology. The theoretical analysis reveals how interaction stiffness non-linearly constrains the system's stability margin, heightening its vulnerability to delay. Furthermore, we validate these theoretical boundaries through experimental trials, highlighting the correlation between analytical stability margins and empirical motor performance. The proposed framework provides rigorous design guidelines for stable remote dyadic systems and suggests the prerequisites for effective delay-compensation strategies.
\end{abstract}

\begin{IEEEkeywords}
Dyadic interaction, time-delayed haptics, bilateral control, motor performance.
\end{IEEEkeywords}

\section{Introduction}
\IEEEPARstart{R}{obot}-mediated human–human (dyadic) interaction enables remote physical cooperation through haptic communication over networks. Such systems have broad applicability, ranging from teleoperation and motor training to remote healthcare and rehabilitation. In rehabilitation robotics, this form of interaction is particularly valuable for preserving the therapeutic benefits of haptic communication between less-skilled patients and more-skilled therapists. This interaction between participants with different skill levels (e.g., experts and novices \cite{DuCampolo2024}, or superior and inferior groups \cite{TakagiBurdet2019}) has been characterised as cooperation \cite{JarrasseBurdet2012}. Previous studies have highlighted the benefits of cooperation in coordinating daily activities \cite{TakagiBurdet2018}, demonstrated its promise for remote healthcare applications \cite{NovakRiener2014}, and shown advantages over individual task performance \cite{GaneshBurdet2014}. Interaction controllers are extensively used in unilateral human-robot systems to ensure compliance, predict motion intention, and maintain smooth movement \cite{ZhouSong2021, LiGe2014}. However, the control dynamics underlying remote dyadic cooperation remain insufficiently understood. Specifically, the relationship between inevitable network time delays and the quality of this physical cooperation requires deeper investigation. While existing research has demonstrated that short time delays do not hinder interaction \cite{IvanovaBurdet2021}, and increased delays adversely affect cooperation by reducing motor performance \cite{DuCampolo2024}, current studies lack systematic approaches to maintaining both stability and interaction efficiency for dyads under delayed conditions.

Through robot mediation, specific parameters can be introduced and controlled to examine their individual effects on dyadic interaction, such as investigating the impact of delay. Previous research has explored the negative impact of network delay on human task performance \cite{AlhalabiKunifuji2003}, yet different types of delay (e.g., visual or haptic) have not been fully distinguished. While delayed haptic feedback has been shown to affect interactive performance adversely \cite{JayHubbold2007}, previous studies often did not account for differences in participants’ skill levels. Ivanova et al. \cite{IvanovaBurdet2021} examined haptic delays in human interaction with a superior human-like robot, concluding that short delays do not significantly deteriorate performance. Additionally, Du et al. \cite{DuCampolo2024} empirically investigated haptic delay effects on human participants with assigned skill levels. However, a rigorous theoretical analysis correlating dyadic controller parameters with interaction stability limits remains unaddressed. Furthermore, it is unclear how these specific limits dictate motor performance.

This study conducts a rigorous stability analysis of robot-mediated dyadic interactions influenced by haptic delay. We utilise the Articares H-MAN, a commercial upper-limb rehabilitation robot, as the experimental platform. Rather than relying on classical bounding inequalities or linear approximations of transcendental delay dynamics, we formulate a dynamic model based on the H-MAN's identified parameters. This model derives exact, closed-form stability boundaries. Consequently, this non-conservative framework explicitly quantifies the multi-variable tradeoff between haptic delay magnitude, interactive stiffness, and inherent robotic dynamics (inertia and damping).

By explicitly delineating this parametric interdependence, the derived criteria function as an analytically buffer mechanism. This exact formulation permits the precise attenuation of interactive stiffness to stabilise the system across multiple delays. Consequently, it guarantees safe cooperation without the transparency degradation typical of conservative controllers. To translate this theoretical framework into an empirical context, we first validate the boundaries through physical experiments. Two H-MAN devices are connected to verify the hardware's fundamental stability limits, independent of human voluntary dynamics.

Subsequently, this research investigates how these theoretical limits functionally impact cooperative task execution. We systematically cross the analytically derived stability thresholds during emulated dyadic trials. This establishes a direct empirical correlation between mathematically defined interaction instability and the tangible deterioration of dyadic motor performance. Ultimately, this work provides a non-conservative mathematical foundation for exact delay mitigation while demonstrating the critical necessity of these boundaries to preserve the efficacy of robot-mediated cooperation.

\section{Problem Formulation and Review of Stability Criteria}
\subsection{Problem Statement}
\noindent A primary objective of this study is to elucidate the correlation between the theoretical stability boundaries of a robot-mediated dyadic system and the resulting human motor performance. In remote haptic interactions, maintaining system stability is the absolute prerequisite to ensure user safety and preserve task efficacy. However, prior to the effective synthesis of buffer functions or active delay-compensation algorithms, one must rigorously define exact stability boundaries dictated by network delay, virtual coupling stiffness, and robotic mediator mechanics. Consequently, the selection of an appropriate mathematical framework to derive these boundaries is critical, as overly conservative estimations can severely degrade the transparency and utility of the haptic interaction.

\subsection{Limitations of Classical Stability Frameworks}
\noindent In the domain of bilateral teleoperation and haptic rendering, time-delay stability has classically been addressed using passivity theory (e.g., scattering approaches, wave variables) \cite{AndersonSpong1989, NiemeyerSlotine1991} or frequency-domain margins (e.g., Llewellyn’s absolute stability criteria) \cite{AdamsHannaford1999}. While passivity-based frameworks guarantee stability under arbitrary unknown passive environments, they are notoriously conservative. By treating the human operator as a worst-case passive environment rather than an active participant, these methods heavily damp the interaction and often degrade haptic transparency \cite{HokayemSpong2006}. Furthermore, classical teleoperation models typically assume a unidirectional master-slave paradigm, which does not accurately represent the asymmetric, bi-directional coupling between two active agents with varying skill levels.

Other classical mathematical techniques similarly present limitations when applied to the extraction of explicit parametric boundaries for precision haptic robotics. In the time domain, Lyapunov-Krasovskii functionals (LKF) \cite{HeLiu2004, KharitonovKharitonov2013} rely on mathematical bounding techniques (such as Jensen’s inequality \cite{ParkPark1999,KimKim2016} or Wirtinger-based ineuqality \cite{SeuretGouaisbaut2013}) to construct tractable linear matrix inequalities. Consequently, LKF methods yield sufficient but not necessary stability conditions, resulting in calculated stability margins that are significantly smaller than the true physical capabilities of the system. In the frequency domain, the Padé approximation is frequently utilised to convert an infinite-dimensional exponential delay term into a rational linear system \cite{RichardRichard2003}. However, it introduces truncation errors that degrade rapidly at higher frequencies, which is critical for crisp haptic rendering. Finally, robust synthesis frameworks such as $\mathcal{H}_\infty$ optimisation \cite{FridmanShaked2002,KayacanKayacan2017} are fundamentally designed to synthesise stabilising controllers subject to bounded uncertainties, rather than to analytically extract precise algebraic stability boundaries for the inherent dynamic parameters of the plant.

\subsection{Zero-Crossing Methodology}
\noindent Applying overly conservative passivity bounds or mathematical approximations to dyadic systems suppresses the very haptic cues necessary for effective cooperation. Therefore, this study leverages an exact zero-crossing frequency-domain approach \cite{GuChen2003}, rather than re-deriving passivity bounds, to establish explicit, closed-form stability boundaries specifically tailored for symmetric mass-spring-damper interconnections. 
\begin{definition}
System instability is identified by the first contact or crossing of the characteristic roots from the stable left-half plane to the unstable right-half plane. By substituting $z = e^{-\delta s}$, the transcendental function is transformed into a bivariate polynomial, yielding $a(s,z)$ and its conjugate polynomial $\bar{a}(s,z) =  z^qa(-s,z^{-1})$, where $q$ is selected to clear any fractional terms. The simultaneous polynomial equations evaluated at the crossing boundary are:
\begin{equation}
    a(s,z) = 0, \quad \bar{a}(s,z) = 0.
    \label{eq: polynomial}
\end{equation} 
Because the characteristic roots appear in conjugate pairs, it suffices to consider zero crossings strictly at positive frequencies. Furthermore, a root at $\omega = 0$ indicates delay-independent marginal stability. Therefore, the system is considered stable against time delay when no positive real frequency $\omega \in \mathbb{R}^+$ satisfies the equilibrium of \eqref{eq: polynomial} for $s = j\omega$. Conversely, delay-induced instability occurs when such an $\omega \in \mathbb{R}^+$ exists, and the maximum tolerable delay $\delta_m$ is determined as:
\begin{equation}
    \delta_m \equiv \min \{\delta \geq 0 \mid \exists \, \omega \in \mathbb{R}^+, \ a(j\omega, e^{-j\delta\omega}) = 0 \}.
\end{equation}
\label{df:zero-crossing}
\end{definition}
Unlike Padé approximations, the zero-crossing approach directly handles the transcendental nature of the delay without truncation errors. Unlike Lyapunov-based inequalities or passivity frameworks, this method yields necessary and sufficient conditions. This approach defines precise, less conservative stability regions that enable highly transparent interactions. By identifying the exact frequency where system roots cross the imaginary axis, we derive an explicit analytical boundary. This formulation maximises the allowable interactive stiffness and delay for the interacting dyad.

\subsection{System Modelling of Robot-Mediated Dyadic Interaction}
\begin{figure}[!t]
\centering
\vspace{0.4em} 
\includegraphics[width=0.7\columnwidth]{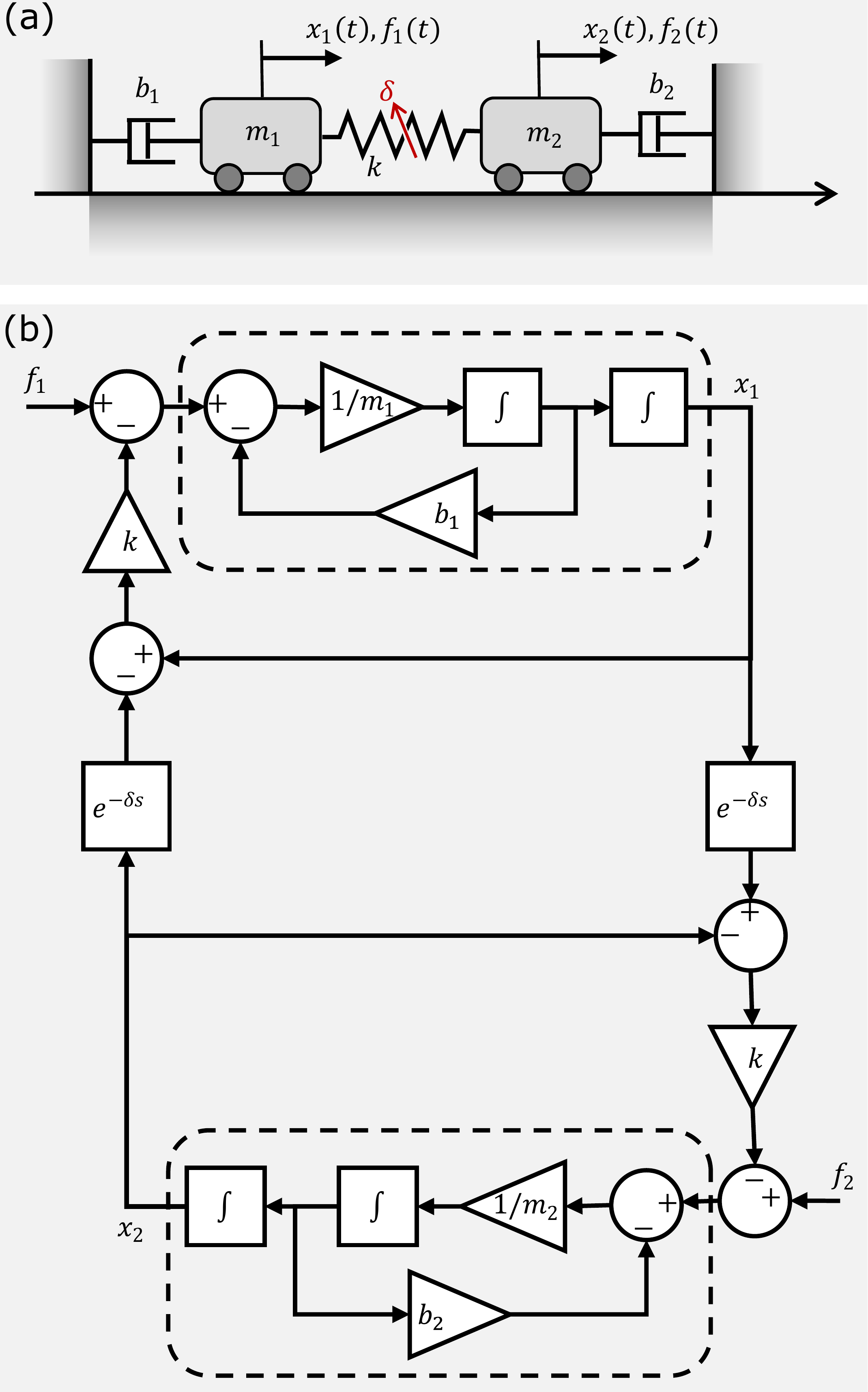}
\caption{(a) Free body diagram of robot-mediated dyadic interactions, modelled as a dyadic mass–spring–damper system. $m_1$ and $m_2$ in kg denote the mass (inertia) of each robot. $b_1$ and $b_2$ in Ns/m denote the damping (friction) of each robot. $k$ (N/m) represents the virtual spring connection. $x_1(t)$ and $x_2(t)$ define the continuous movement in metres. $f_1(t)$ and $f_2(t)$ define the interactive force in N exerted by human operators on the robots. (b) Control diagram of dyadic interaction with round-trip time delay. The time delay $\delta$ is expressed in seconds. The dashed box highlights the dynamic system of each robotic mediator.}
\label{fig:fbd}
\end{figure}
\noindent To investigate the system stability of robot-mediated dyadic interaction, we establish a physical model based on the Articares H-MAN. While clinical robotic systems often exhibit complex multi-DOF configurations, the adoption of a decoupled mass-spring-damper model is a well-established and robust abstraction in the field of neurorehabilitation and haptic manipulation \cite{HoganHogan1984, KrebsVolpe1998}. Foundationally, rehabilitation robots are designed as high-performance impedance interfaces where mechanical properties such as transparency are prioritised \cite{CampoloMasia2014}. Under such conditions, the dominant dynamics of the system can be accurately represented by second-order linear models, an assumption verified across several landmark platforms including the MIT-MANUS \cite{KrebsVolpe1998} and various commercial haptic interfaces \cite{AdamsHannaford1999}. 

In this model, illustrated in Fig.~\ref{fig:fbd}, the dynamic parameters of the two robotic mediators are defined by their respective masses, $m_1$ and $m_2$, and their viscous damping coefficients, $b_1$ and $b_2$. This linear system modelling has been addressed in the study of H-MAN as a rehabilitation platform \cite{CampoloMasia2014}. The virtual haptic coupling connecting the two users across the network is represented as a virtual spring with stiffness $k$. This elastic coupling inherently functions as a fundamental impedance controller, a paradigm widely adopted to regulate compliant physical human-robot interactions \cite{LiGe2014,YangLi2018}. The continuous position trajectories of the robots are denoted by $x_1(t)$ and $x_2(t)$, while $f_1(t)$ and $f_2(t)$ represent the exogenous interactive forces exerted by the human operators. Under the assumption of a constant, symmetric network-induced communication delay $\delta$ corresponding to each directional channel, the continuous-time dynamic equations of motion for the coupled dyadic system are formulated as:
\begin{equation}
\begin{split}
    m_1\ddot{x}_1(t) + b_1\dot{x}_1(t) &= k(x_2(t-\delta_1) - x_1(t)) + f_1(t), \\
    m_2\ddot{x}_2(t) + b_2\dot{x}_2(t) &= k(x_1(t-\delta_2) - x_2(t)) + f_2(t).
\end{split}
\label{eq:time_domain_model}
\end{equation}
Equation \eqref{eq:time_domain_model} forms the specific plant dynamics of the dyadic interaction. The subsequent section transforms this model into a generalised frequency-domain framework to extract the exact algebraic stability boundaries.

\section{Analytical Derivation of Exact Stability Boundaries for Delayed Dyadic Interactions}
\subsection{Generalised Transfer Function Architecture}
\begin{figure}[!t]
    \centering
    \includegraphics[width=\columnwidth]{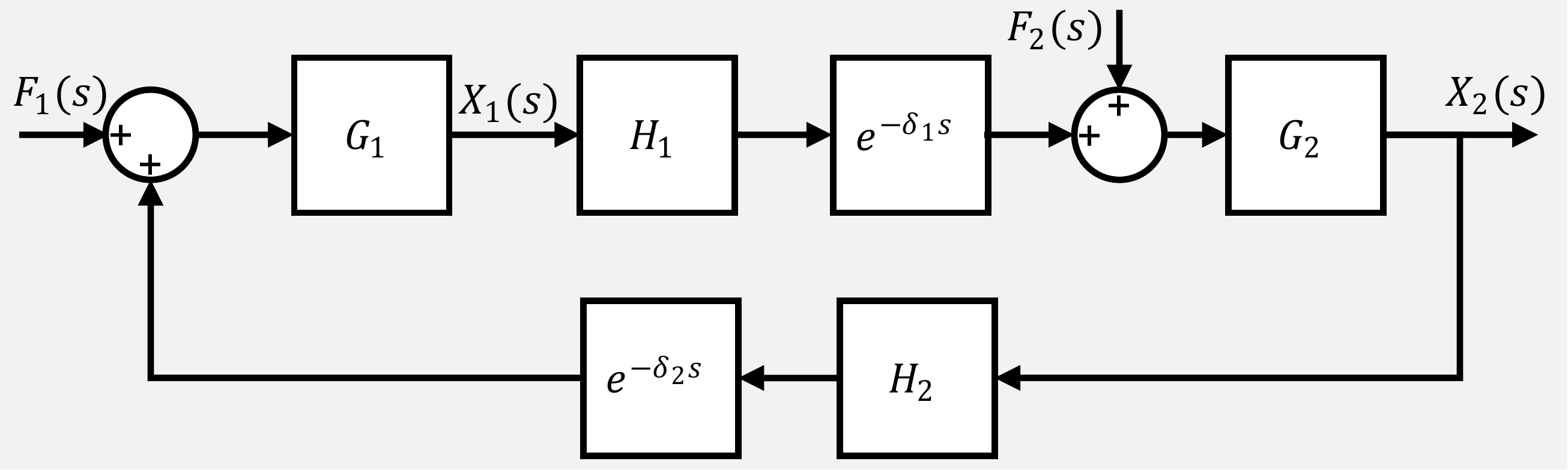}
    \caption{Generalised transfer function block diagram of the robot-mediated dyadic interaction system. The blocks $G_1(s)$ and $G_2(s)$ represent the arbitrary plant dynamics of the individual robotic interfaces. The blocks $H_1(s)$ and $H_2(s)$ denote the haptic communication networks transmitting the coupling forces. The parameters $\delta_1$ and $\delta_2$ explicitly define the asymmetric network-induced time delays corresponding to the forward and return communication channels, respectively.}
    \label{fig: tf}
\end{figure}

\noindent 
The control paradigm for dyadic robot mediation can be represented via a generalised closed-loop transfer function model, as illustrated in Fig.~\ref{fig: tf}. In this generalised architecture, $G_1(s)$ and $G_2(s)$ represent the dynamic plant models of the two multi-degree-of-freedom (DOF) robotic interfaces. The haptic coupling transmitted over the network is denoted by the feedback communication channels $H_1(s)$ and $H_2(s)$, which inherently incorporate the directional network-induced time delays, yielding the delayed haptic signals $e^{-\delta_1 s}$ and $e^{-\delta_2 s}$.   
Based on this architecture, the closed-loop transfer function matrix in the frequency domain is derived as:
\begin{equation}
\begin{split}
    \frac{\partial \mathbf{X}}{\partial \mathbf{F}} = \frac{1}{P(s)}\begin{bmatrix}
       G_1 & G_1G_2H_1e^{-\delta_1s}\\
       G_1G_2H_2e^{-\delta_2s} & G_2
    \end{bmatrix},
\end{split}
\end{equation}
the characteristic polynomial of the system is defined by the denominator:
\begin{equation}
    P(s) = 1 - G_1(s)G_2(s)H_1(s)H_2(s)e^{-(\delta_1 + \delta_2)s} = 0.
\end{equation}
This characteristic equation is transcendental due to the exponential delay term. As demonstrated by Gu et al. \cite{GuChen2003}, such systems can be resolved using the zero-crossing method (see Definition \ref{df:zero-crossing}).

While classical stability margins heavily rely on symmetric delay assumptions, a distinct novelty of this work is the expansion of the zero-crossing methodology to be fully adaptive to bilateral asymmetric interactions. By defining a lumped delay variable $z = e^{-(\delta_1 + \delta_2)s}$, we transform the generalised characteristic equation into a bivariate polynomial $P(s,z)$. The polynomial and its conjugate form $\bar{P}(s,z) = zP(-s,z^{-1})$ can be explicitly written as:
\begin{equation}
\begin{split}
    P(s,z) &= 1 - G_1 G_2 H_1 H_2 z = 0, \\
    \bar{P}(-s,z^{-1}) &= z - \bar{G}_1 \bar{G}_2 \bar{H}_1 \bar{H}_2 = 0.
\end{split}
\end{equation}
By solving the forward polynomial $P(s,z) = 0$ for $z$, we uniquely isolate the delay parameter $z(\omega)$ as a function of the plant and controller dynamics:
\begin{equation}
    z(\omega) = \left. \frac{1}{G_1 G_2 H_1 H_2} \right|_{s=j\omega}.
    \label{eq:z_omega}
\end{equation}
A highly advantageous property of this formulation emerges when computing the crossing frequency $\omega$. By substituting $s = j\omega$ and incorporating $z(\omega)$ into the conjugate equation $\bar{P}(s,z) = 0$, the delay parameter $z$ is algebraically eliminated. This yields a purely frequency-dependent magnitude condition:
\begin{equation}
    \left. 1 - G_1 G_2 \bar{G}_1 \bar{G}_2 H_1 H_2 \bar{H}_1 \bar{H}_2 \right|_{s=j\omega} = 0.
    \label{eq:magnitude_condition}
\end{equation}
Using Equation \eqref{eq:magnitude_condition}, the critical crossing frequency $\omega \in \mathbb{R}$ is derived independently of the delay magnitude. Once $\omega$ is obtained, Euler's formula can be leveraged to represent the complex delay boundary as $z(\omega) = e^{j\operatorname{arg}\left[z(\omega)\right]}$. Given that $z = e^{-(\delta_1 + \delta_2) s}$ and $s = j\omega$, the maximum tolerable delay for the asymmetric dyadic interaction is formulated as:
\begin{equation}
    (\delta_1 + \delta_2)_{\operatorname{max}}   = -\frac{1}{\omega}\arg\left[z(\omega)\right],
    \label{eq:delta_begin}
\end{equation}
where $\arg(z)$ computes the principal argument of the complex variable. This generalised algebraic decoupling forms a robust, exact boundary for assessing the stability of heterogeneous, robot-mediated cooperative systems. 

\subsection{Explicit Stability Boundaries for Heterogeneous Dyadic Systems}
\noindent To derive explicit analytical stability boundaries for the heterogeneous dyadic interaction modelled in \eqref{eq:time_domain_model}, the plant dynamics are specifically instantiated as $G_i(s) = 1/(m_is^2 + b_is + k)$ for $i \in \{1, 2\}$, where $m_i$ and $b_i$ represent the identified inertia and damping parameters of each respective robotic platform. Under the assumption of a symmetric virtual elastic coupling with stiffness $k$, the haptic communication channels are defined as $H_1(s) = H_2(s) = k$. 

Substituting these specific transfer functions into the generalised magnitude condition \eqref{eq:magnitude_condition} yields the governing frequency-domain equation for the coupled system. Expanding the resultant expression reveals that the constant $k^4$ terms perfectly cancel, allowing an $\omega^2$ factor to be extracted from the original 8th-order polynomial. As established in Definition \ref{df:zero-crossing}, the root at $\omega = 0$ represents delay-independent marginal stability (a rigid-body mode). Therefore, the delay-induced oscillatory instability is strictly governed by the remaining 6th-order polynomial, $\mathcal{F}(\omega)$. Solving for the critical crossing frequency requires finding an $\omega$ such that:
\begin{equation}
    \omega \in \mathbb{R}^+ \;\land\; \mathcal{F}(\omega) = 0,
\end{equation}
where the frequency polynomial $\mathcal{F}(\omega)$ is expanded as:
{\small\begin{equation}
\begin{split}
    \mathcal{F}(\omega) &= m_1^2 m_2^2 \omega^6 \\
    &+ \left[(b_1^2 - 2 k m_1) m_2^2 + (b_2^2 - 2 k m_2) m_1^2 \right] \omega^4 \\
    &+ \left[(b_1^2 - 2 k m_1)(b_2^2 - 2 k m_2) + k^2 (m_1^2 + m_2^2) \right] \omega^2 \\
    &+ k^2 \left[(b_1^2 - 2 k m_1) + (b_2^2 - 2 k m_2)\right].
    \label{eq:f_omega}
\end{split}
\end{equation}}
Once a valid crossing frequency $\omega$ is identified, it is substituted back into the delay parameter expression $z(\omega)$ initially defined in \eqref{eq:z_omega}. Evaluating this expression at $s = j\omega$ explicitly isolates the complex delay boundary:
\begin{equation}
    z(\omega) = \frac{1}{k^2}(-m_1\omega^2 + j b_1\omega + k)(-m_2\omega^2 + j b_2\omega + k).
    \label{eq:delta_end}
\end{equation}
For practical dyadic implementations, the communication channels are typically subject to symmetric one-way network delays ($\delta_1 = \delta_2 = \delta$). Under this condition, the maximum tolerable one-way delay $\delta_{m}$ can be directly computed from the phase of $z(\omega)$ derived from \eqref{eq:delta_begin} as:
\begin{equation}
    \delta_{m} \equiv \mathcal{D}(m_1,b_1,m_2,b_2,k) = -\frac{1}{2\omega}\arg\left[z(\omega)\right].
\end{equation}
Crucially, if $k$ is sufficiently small, the interaction achieves \textit{delay-independent stability}, meaning no real roots exist for $\mathcal{F}(\omega)$. For this 6th-order direct coupling architecture, the absolute maximum delay-independent stiffness ($k_{m}$) can be extracted analytically by applying Descartes' Rule of Signs to constrain the polynomial coefficients. The rigorous algebraic proof formulating this exact critical stiffness limit is detailed in Appendix \ref{appendix: coupling}.

\subsection{Stability Analysis and Effect of Model-based Parameters}
\begin{figure}[!t]
\centering
\includegraphics[width=\columnwidth]{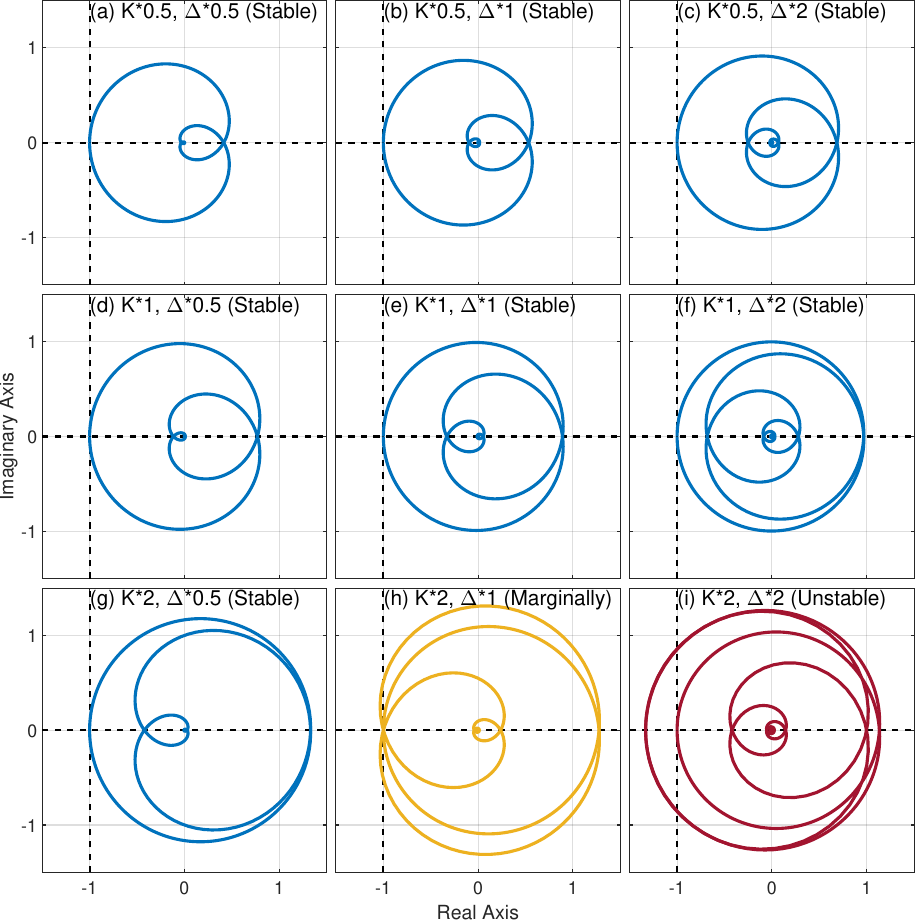}
\caption{The poles of the open-loop transfer function, obtained from \eqref{eq:time_domain_model}, are strictly negative. The Nyquist criterion is applied to assess system stability by examining the encirclement of the point $(-1,0)$. To cover a sufficient range of controllable parameters, stiffness is varied from $0.5\K$, $\K$, to $2\K$, where $\K = \mathcal{S}(\M,\B,\M,\B)$, and the time delay is varied from $0.5\Delta$, $\Delta$, to $2\Delta$, where $\Delta = \mathcal{D}\left(\M,\B,\M,\B,2\K\right)$. The nominal values of $\M$, $\B$, $\K$, and $\Delta$ are summarised in Table \ref{tab:notation}.}
\label{fig:nyquist}
\end{figure}

\noindent The maximum tolerable delay $\delta_m$ can be computed from \eqref{eq:delta_begin}–\ref{eq:delta_end}. For ideal dyadic interaction mediation by robots, it can be assumed that both robots possess identical dynamic parameters, i.e. $m = m_1 = m_2$ and $b = b_1 = b_2$. The equations above indicate that the system is delay-independent stable when $k \leq k_m$, where $k_m = \frac{b^2}{2 m}$. When $k > k_m$, the system remains stable if the time delay satisfies $\delta < \delta_m$, where
\begin{equation}
    \delta_m = \frac{-m}{\sqrt{2 m k - b^2}} \operatorname{arg}\left[\frac{m k - b^2}{m k} - j \frac{b \sqrt{2 m k - b^2}}{m k}\right].
    \label{eq:ideal}
\end{equation}
Thus, for identical dyadic robots, the maximum tolerable delay $\delta_m$ can be directly obtained from \ref{eq:ideal}. Assuming an ideal model ($m_1 = m_2 = \M$, $b_1 = b_2 = \B$), we study the stability and the correlation among different parameters. From \eqref{eq:delta_begin}–\eqref{eq:delta_end}, we can formulate the following conjectures:
\begin{enumerate}
    \item The system is delay-independent stable when $k \leq k_m$.
    \item The system is stable when $k > k_m$ and $\delta \leq \delta_m$.
    \item The system is unstable when $k > k_m$ and $\delta > \delta_m$.
\end{enumerate}
The Nyquist criterion was plotted to verify the conjectures using different stiffness and delay values, as shown in Fig.~\ref{fig:nyquist}. The interactive stiffness was set to $0.5\K$, $\K$, and $2\K$, where $\K = \mathcal{S}(\M,\B,\M,\B)$. The time delay was set to $0.5\Delta$, $\Delta$, and $2\Delta$, where $\Delta = \mathcal{D}(\M,\B,\M,\B,2\K)$. Because no feasible finite solution of $\mathcal{D}$ exists when $k \leq \K$, $\Delta$ was selected by substituting $2\K$. From the Nyquist plots, only Fig.~\ref{fig:nyquist}(i) exhibits system instability. Therefore, the system is delay-independent stable when the stiffness is $\K$ or $0.5\K$. When the stiffness is set to $2\K$, the system is stable when the delay is $0.5\Delta$ or $\Delta$, but becomes unstable when the delay is $2\Delta$. It is noted that the system is marginally stable when the stiffness is $2\K$ and the delay is $\Delta$. These results are consistent with our conjectures; however, this Nyquist-based stability analysis does not reveal the effect of inertia and damping on the system's vulnerability.

\begin{figure*}[!t]
    \centering
    \vspace{0.4em} 
    \includegraphics[width=\textwidth]{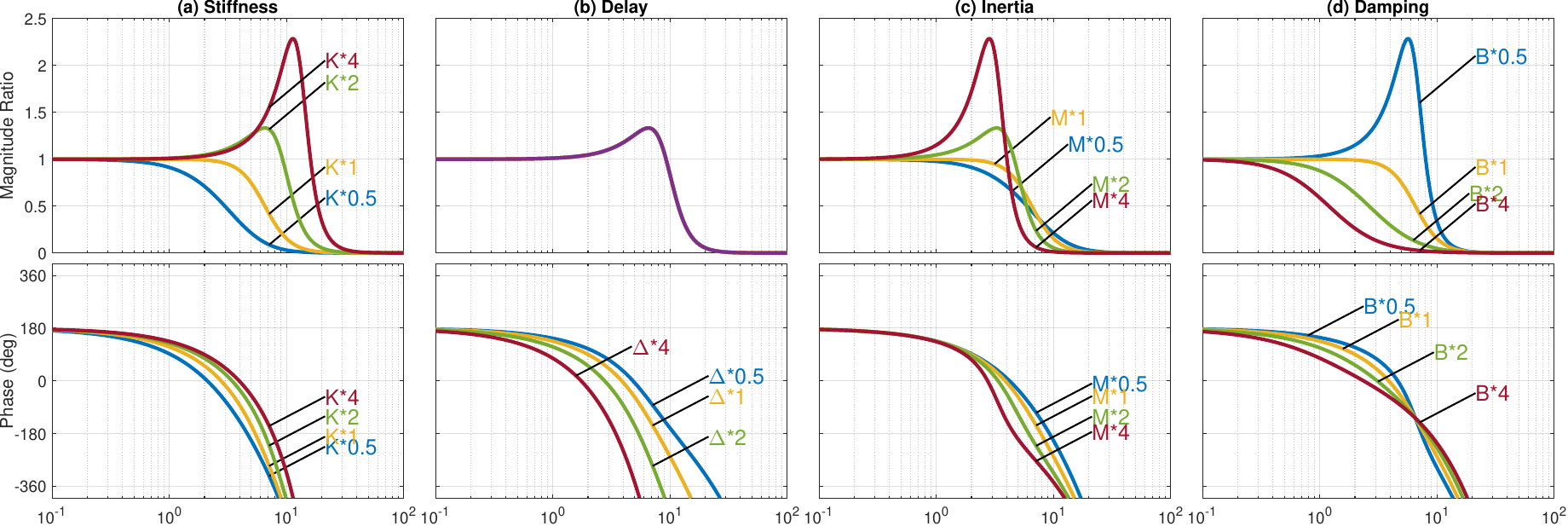}
    \caption{Effect of modelled stiffness, delay, inertia, and damping on the magnitude ratio and phase shift, shown as the frequency response of the delayed dyadic mass–spring–damper system. Magnitude ratio has been used to indicate correlations among different parameters of a multi-DOF mass–spring–damper system \cite{DavidsonCharles2017}. Phase (in degrees) indicates how the frequency response is shifted or delayed relative to the input. (a) Increasing stiffness alone increases the magnitude ratio at higher input frequencies. Stiffness greater than $\mathrm{K}$ ($2\mathrm{K}$ or $4\mathrm{K}$) can increase the magnitude ratio above $1$, which may cause the response to diverge depending on the frequency. (b) Increasing delay alone shifts the phase to the left, delaying the response without affecting the magnitude ratio. (c) Increasing inertia usually increases the magnitude ratio, thereby reducing stability under frequencies near $-180$\,deg phase shift. (d) Increasing damping enhances stability by reducing the magnitude ratio; the system becomes unstable when damping is less than $\B$ (e.g., $0.5\B$).}
    \label{fig:bode}
\end{figure*}

From \eqref{eq:delta_begin}–\eqref{eq:delta_end}, the correlation between the maximum tolerable delay and interactive stiffness is directly influenced by the dynamic parameters of the robots, namely inertia and damping. Magnitude ratio plots have previously been employed to study the effects of inertia and viscoelastic properties on robotic systems \cite{DavidsonCharles2017}. Fig.~\ref{fig:bode} illustrates both the magnitude ratio and phase shift of the frequency response as affected by stiffness, delay, inertia, and damping. As shown in Fig.~\ref{fig:bode}(b), increasing delay induces a negative phase shift, thereby reducing stability. Fig.~\ref{fig:bode}(a) demonstrates that the system becomes progressively unstable as stiffness increases beyond $\K$. The system also exhibits greater vulnerability with increasing inertia (Fig.~\ref{fig:bode}(c)) and with decreasing damping (Fig.~\ref{fig:bode}(d)). Since the critical stiffness formulated by $\mathcal{S}$ is determined by the dynamic parameters, the stiffness at which the system becomes vulnerable to time delay is determined by the estimation of the dynamic identifications. In our mass–spring–damper system, the delay-independent criterion is particularly sensitive to the damping coefficients due to their quadratic dependence in \eqref{eq:delta_begin}.
\begin{figure}[!t]
\centering
\includegraphics[width=\columnwidth]{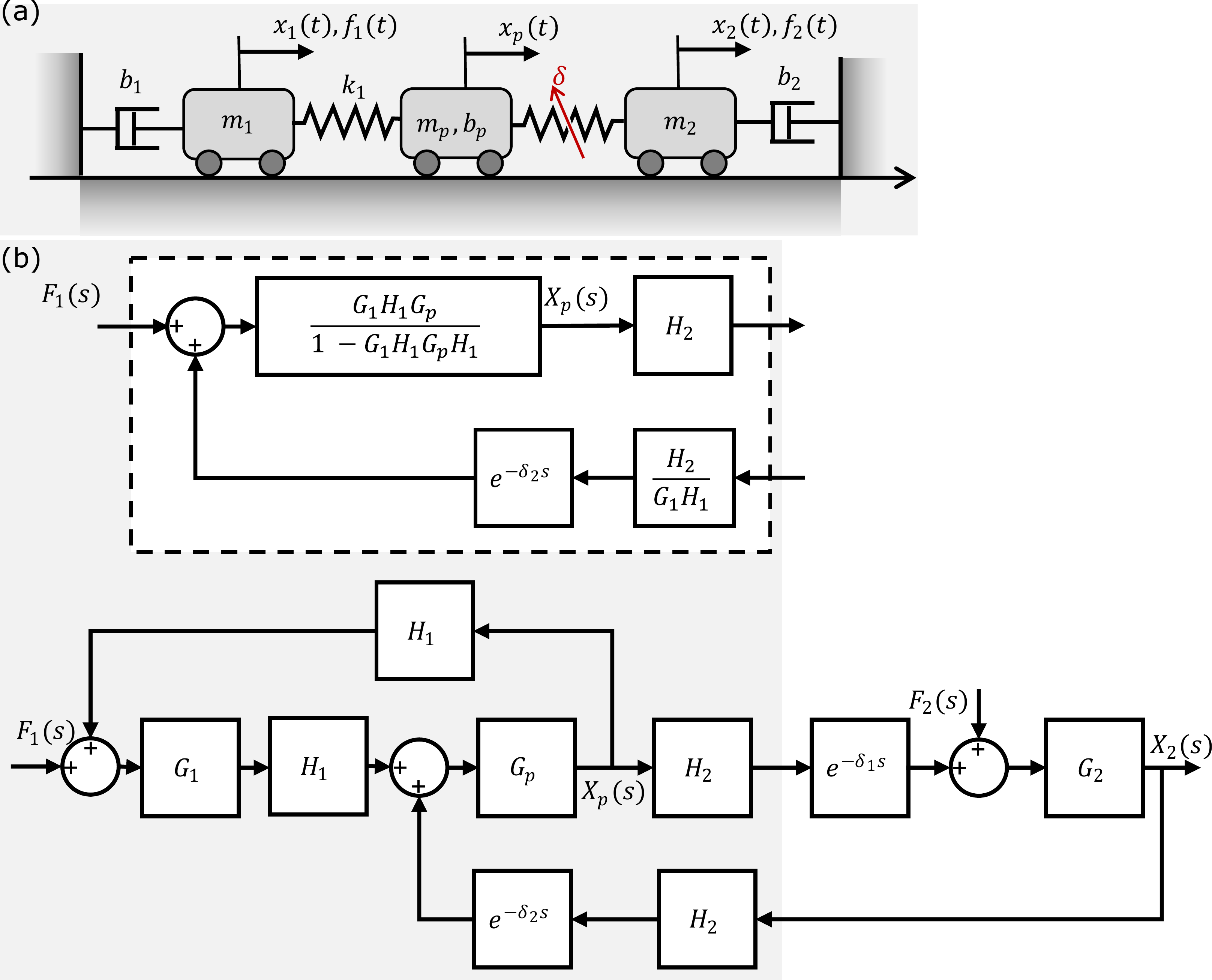}
\caption{(a) Free body diagram of the proxy-mediated dyadic interaction, represented via mechanical mass-spring-damper elements. The virtual proxy ($m_p$, $b_p$) is co-located with Robot$\mathrm{_1}$. The delay $\delta$ only affects the transmission between the proxy and Robot$\mathrm{_2}$. (b) Transfer function block diagram for the virtual proxy dyadic interaction. The internal proxy dynamics are structurally absorbed into the generalised forward and return transfer functions, enabling the application of the universal zero-crossing criterion.}
\label{fig:proxy_fbd}
\end{figure}
\subsection{Analytical Extension to Virtual Proxy Architectures}
\noindent While the direct elastic coupling establishes a fundamental baseline, certain cooperative paradigms expand the network topology to include a shared virtual proxy \cite{KagerCampolo2019}. Basdogan et al. demonstrated that utilising a shared dynamic object as a medium for haptic communication significantly enhances dyadic task execution and the operators' sense of togetherness \cite{BasdoganSlater2000}. In this architecture, human operators manipulate this central interactive medium, which is governed by its own independent dynamic properties (mass $m_p$ and damping $b_p$). Assuming the proxy is co-located on the local server of Robot$_\mathrm{1}$ with a direct coupling stiffness $k_1$, the remote Robot$_\mathrm{2}$ interacts with this shared proxy over the delayed network via stiffness $k_2$. The resulting time-domain equations of motion are formulated as:
{
\small
\begin{equation}
    \begin{split}
        m_1\ddot{x}_1 + b_1\dot{x}_1 &= k_1(x_p(t) - x_1(t)) + f_1(t)\\
        m_p\ddot{x}_p + b_p\dot{x}_p &= k_1(x_1(t) - x_p(t))+ k_2(x_2(t-\delta) - x_p(t))\\
        m_2\ddot{x}_2 + b_2\dot{x}_2 &= k_2(x_p(t-\delta) - x_2(t)) + f_2(t)
    \end{split}
\label{eq:proxy_model}
\end{equation}
}
We introduce this virtual proxy topology strictly to demonstrate the analytical scalability of the proposed zero-crossing framework. By evaluating this model, we prove that our boundary extraction methodology seamlessly adapts to highly complex, higher-order asymmetric topologies without resorting to conservative approximations.

By taking the Laplace transform, the proxy subsystem is mapped into the generalised frequency-domain architecture defined in Fig.~\ref{fig: tf}. The equivalent plant and communication transfer functions are algebraically identified as:
{\small	
\begin{equation}
    \begin{split}
        G_1'(s) &= \frac{k_1}{(m_ps^2 + b_ps + k_1 + k_2)(m_1s^2 + b_1s + k_1) - k_1^2}\\
        H_1'(s) &= k_1\\
        G_2'(s) &= (m_2s^2 + b_2s + k_2)^{-1}\\
        H_2'(s) &= k_2k_1^{-1}(m_1s^2 + b_1s + k_1) 
    \end{split}
\end{equation}
}Because these constituent transfer functions inherently encapsulate the proxy dynamics, substituting them into the generalised magnitude condition \eqref{eq:magnitude_condition} fundamentally alters the characteristic equation. This expansion produces a highly complex 10th-order frequency polynomial, $\mathcal{F}_{p}(\omega)$, and a correspondingly extended complex delay parameter, $z_{p}(\omega)$. Due to their substantial algebraic length, the full explicit closed-form expansions for both $\mathcal{F}_{p}(\omega)$ and $z_{p}(\omega)$ are strictly catalogued in Appendix \ref{appendix: proxy}. Therefore, this virtual proxy topology validates the analytical scalability of our boundary extraction methodology, proving that it can seamlessly adapt to highly complex, higher-order asymmetric topologies without resorting to conservative approximations.

\section{Stability for empirical dyadic robots}
\subsection{Dynamic Identification}
\noindent H-MAN is a planar robot developed for upper-limb rehabilitation \cite{CampoloMasia2014}. Two H-MANs (H-MAN\textsubscript{1} and H-MAN\textsubscript{2}) served as mediators in the experiments on dyadic interactions. Based on \eqref{eq:delta_begin}–\eqref{eq:delta_end}, the study of the correlation between stiffness and delay depends on the dynamic parameters of the mediated robots, namely mass and damping. System identification is therefore required to obtain the estimates of the dynamic parameters before conducting the experiments. Since the H-MAN is a linear two-DOF robot operating along the $\operatorname{x}$ and $\operatorname{y}$ axes, its dynamic model is assumed by identifying each axis independently.
Using Fourier excitation \cite{SweversVanBrussel1997} and the weighted least squares estimation method \cite{WuYou2010}, we obtained the estimated inertia and damping along each axis for H-MAN\textsubscript{1} ($\mathrm{M_1^{\operatorname{x}}}$, $\mathrm{B_1^{\operatorname{x}}}$, $\mathrm{M_1^{\operatorname{y}}}$, $\mathrm{B_1^{\operatorname{y}}}$) and H-MAN\textsubscript{2} ($\mathrm{M_2^{\operatorname{x}}}$, $\mathrm{B_2^{\operatorname{x}}}$, $\mathrm{M_2^{\operatorname{y}}}$, $\mathrm{B_2^{\operatorname{y}}}$). Therefore, this 2-DOF linear dynamic parameters (for the $\operatorname{x}$ and $\operatorname{y}$ axes) of H-MAN\textsubscript{1} and H-MAN\textsubscript{2} can be estimated, with the resulting values listed in Table \ref{tab:notation}, and the regression of estimated parameters plotted in Fig.~\ref{fig:leastSquare}. The stability criteria for dyadic interactions mediated by a multi-DOF robotic system can be formulated as
\begin{equation}
\begin{split}
    \bar{k}_m &= \min \{k_m^i = \mathcal{S}(m_1^i, b_1^i, m_2^i, b_2^i)\}, \\
    \bar{\delta}_m &= \min \{\delta_m^i = \mathcal{D}(m_1^i, b_1^i, m_2^i, b_2^i, k)\} \operatorname{for}\,k>\bar{k}_m,
    \label{eq:mutiple}
\end{split}
\end{equation}
where $i$ indexes the DOFs of the robotic system ($i = x, y$ in our system). $\bar{k}_m$ represents the critical stiffness for delay-independent stability, and $\bar{\delta}_m$ represents the maximum tolerable delay for delay-dependent stability in a multi-DOF robotic system. By substituting the known parameter values into \eqref{eq:mutiple}, we can obtain the base values for the controlled experiment: the base stiffness 
$
\KXY = \min_{i=x,y} \left\{ \mathcal{S}(\mathrm{M_1^i, B_1^i, M_2^i, B_2^i}) \right\},
$
and the base delay by introducing a larger stiffness $2\KXY$, yielding
$
\mathrm{\DXY} = \min_{i=x,y} \left\{ \mathcal{D}(\mathrm{M_1^i, B_1^i, M_2^i, B_2^i, 2 \KXY}) \right\}.
$ 

\subsection{Experimental Validation of Analytical Stability}
\noindent The stability validation was specifically designed to isolate the system's inherent stability properties from human. This approach establishes a robust empirical baseline, as these exact conditions are subsequently utilised as the reference framework to evaluate dyadic motor performance. To systematically sample across the theoretical stability boundary, the interactive stiffness $k$ was configured at four proportional levels: $0.5\KXY$, $\KXY$, $2\KXY$, and $4\KXY$. For each stiffness configuration, the network delay was tested at four discrete magnitudes: $0$, $0.5\DXY$, $\DXY$, and $2\DXY$. During each trial, the robots were commanded to hold a fixed equilibrium position offset along a 45-degree diagonal through the workspace centre, at $(0.1, 0.1)$\,m and $(-0.1, -0.1)$\,m relative to the origin. This created an initial Euclidean displacement of approximately $0.28$\,m. By enforcing this initial displacement, the system generates a steady-state restoring force that simultaneously stresses both axes, allowing the unforced transient and steady-state responses to be clearly observed under delayed coupling.

The system response for each specific stiffness-delay condition was recorded for 30 seconds. In experimental haptic systems, instability typically manifests as sustained limit-cycle oscillations or high-frequency chatter driven by delayed discrete-time rendering and unmodelled hardware dynamics \cite{DiolaitiSalisbury2006}. Therefore, stability was evaluated post-hoc using the standard deviation of the position signal in the steady-state window (the final 10 seconds of the trial). A trial was strictly classified as unstable if the standard deviation exceeded a 0.01 m noise threshold (indicating the onset of sustained limit cycles), or if divergent high-frequency chatter caused the hardware to trigger fundamental safety limits, specifically by exceeding the defined workspace boundaries. This boundary violation occurs when growing, unstable forces push the end-effector outside of the robot's safe reachable area, prompting an automatic system halt.
\begin{figure}[!t]
    \centering
    \includegraphics[width=\columnwidth]{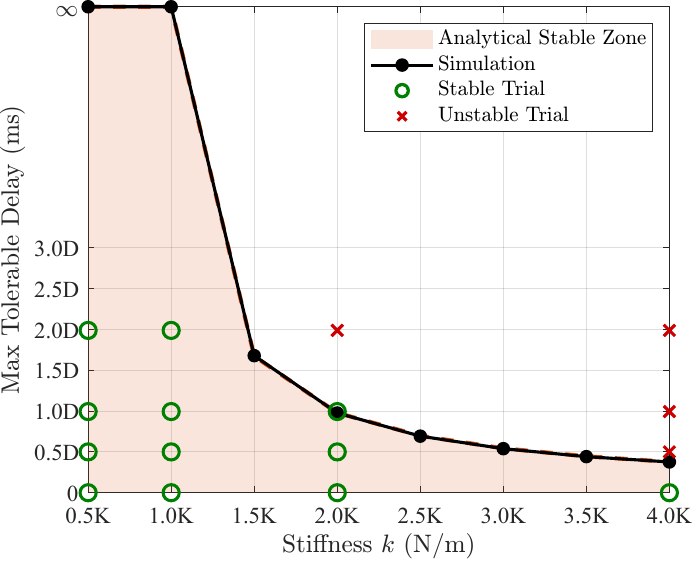}
    \caption{Experimental validation of stability boundaries under constant network-induced delay. The shaded orange region represents the theoretical safe zone computed via the proposed exact zero-crossing criterion. Discrete markers indicate physical experimental trials performed on the H-MAN dyadic setup: green circles ($\bullet$) denote stable trials where the position error remained bounded, while red crosses ($\times$) denote unstable trials exhibiting divergent oscillations or safety trigger activation. The experimental data confirm that the robust analytical criterion provides a highly accurate and valid lower bound, with all unstable trials occurring strictly outside the predicted safe zone.}
    \label{fig:experimental_results}
\end{figure}

The experimental results are presented in Fig.~\ref{fig:experimental_results}. Stable trials are marked with green circles, while unstable trials are marked with red crosses. The experimental data demonstrate strong concordance with the theoretical predictions. The transition from stable to unstable behaviour occurs consistently near the boundary of the analytical stable zone. Crucially, no instability was observed within this analytical stable zone, confirming that the proposed robust zero-crossing criterion provides a reliable and conservative safety margin for physical hardware operation. The alignment between the physical experiment and the theoretical model validates the efficacy of the proposed design guidelines for ensuring stability in real-world networked haptic applications.

\section{Experimental Methodology}
\subsection{Experimental Setup}
\noindent We empirically investigated the impact of delay-induced stability on dyadic interactions mediated by two H-MAN devices. A complementary, torque-sensing robotic system was developed, as shown in Fig.~\ref{fig:setup}(b). This anthropomorphic structure comprises two active HEBI actuators (shoulder and elbow) and one passive bearing (as the wrist joint). The HEBI-based robotic arm serves two critical purposes. First, it provides direct torque measurement to estimate the H-MAN's system dynamics (dynamic identification). Second, it executes programmable trajectories (e.g., sinusoidal motions) to systematically excite H-MAN's dynamics. This approach replicates human-like interaction patterns with repeatability and precision beyond manual manipulation. Prior to conducting experiments, the torque control of the HEBI actuators was calibrated.

\subsection{Task Description}
\begin{figure}[!t]
    \centering
    \vspace{0.4em} 
    \includegraphics[width=\columnwidth]{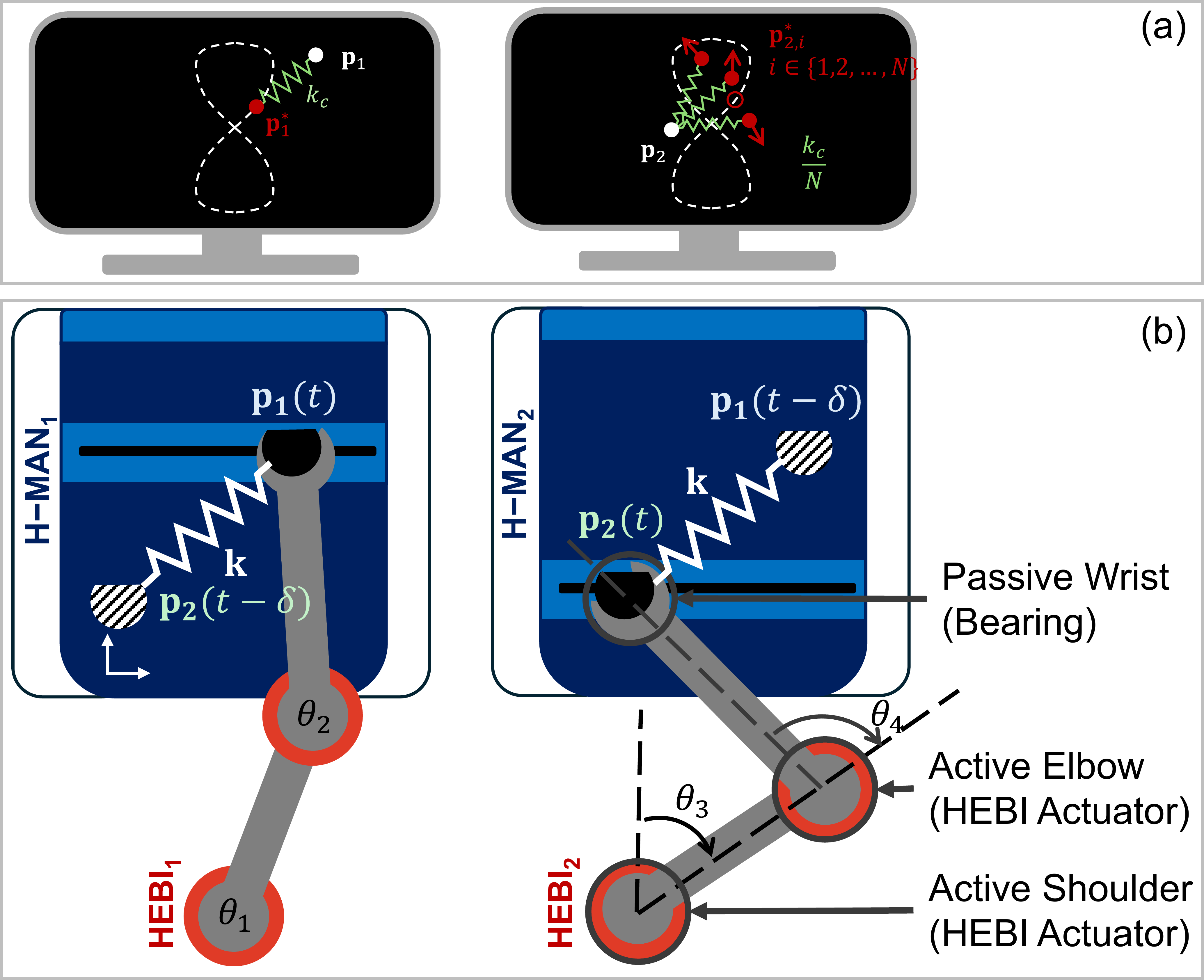}
    \caption{Both H-MANs were operated by an automated robotic system based on HEBI actuators. (a) Haptic uncertainty, mapped from visual disturbance, is introduced to simulate a novice participant interacting with an expert participant. (b) The grippers of H-MANs are physically connected to the end-effector of the HEBI-based robotic system. $\theta_i$ (rad) denotes the angular position from the initial joint configuration. A virtual spring with stiffness $\mathbf{k}$ (N/m), where $\mathbf{k} = [k_x\,\,k_y]^{\mathrm{T}}$, is implemented. $\mathbf{p}_1(t)$ and $\mathbf{p}_2(t)$ (m) denote the real-time positions of H-MAN\textsubscript{1} and H-MAN\textsubscript{2}, respectively, with $\mathbf{p}_i = [p_i^x\,\,p_i^y]^{\mathrm{T}}$. Therefore, $\mathbf{p}_1(t-\delta)$ and $\mathbf{p}_2(t-\delta)$ represent the delayed positions by $\delta$\,s.}
    \label{fig:setup}
\end{figure}
\noindent The experimental task was designed as a tracking exercise. The nominal target trajectory is a time-variant, preprogrammed movement $\mathbf{p}^*(t) = 
\begin{bmatrix} 
    A\sin(2\omega t) &
    B\sin(\omega t) 
\end{bmatrix}^{\mathrm{T}}$, with $A = 0.05$\,m, $B = 0.1$\,m, and $\omega = 2.59\,\mathrm{rad\,s^{-1}}$.  
Turlapati et al.~\cite{TurlapatiCampolo2024} investigated a human-like tracking task using a figure shape identical to that described by $\mathbf{p}^*(t)$. Their collected data in~\cite{TurlapatiCampolo2024} exhibited an angular frequency of $2.59 \pm 0.42\,\mathrm{rad\,s^{-1}}$ ($n=10$). To emulate an expert participant, one HEBI-based robot (denoted HEBI\textsubscript{1}) is connected to H-MAN\textsubscript{1}. Its target trajectory strictly matches the nominal trajectory without disturbance, $\mathbf{p}_{1}^*(t) = \mathbf{p}^*(t)$. The kinematics are defined as:
\begin{equation}
    \boldsymbol{\tau}_1 = \mathbf{J}(\theta_1,\theta_2)^\mathrm{T}\mathbf{F}_1, \quad \mathbf{F}_1 = k_c(\mathbf{p}_{1}^* - \mathbf{p}_{1}),
\end{equation}
where $\boldsymbol{\tau}_1$ denotes the commanded torques for HEBI\textsubscript{1}. The human-like motion is generated through a virtual stiffness set as a constant $k_c = 120\,\mathrm{N\,m^{-1}}$. This approach follows previous studies that used virtual compliant elastic bands to emulate human-like environments \cite{TakagiBurdet2017}. HEBI\textsubscript{2} emulates a novice participant. Its target trajectory is rendered as a blurry cloud of moving points, as shown in Fig.~\ref{fig:setup}(a). The visual disturbance, a method used to manipulate task difficulty levels \cite{TakagiBurdet2019}, is extended in this work by mapping it into a corresponding haptic disturbance. Specifically, the blurred target is represented by multiple moving spots. Each rapidly moving point generates a haptic perturbation, collectively increasing task uncertainty. The corresponding kinematics are:
\begin{equation}
    \boldsymbol{\tau}_2 = \mathbf{J}(\theta_3,\theta_4)^\mathrm{T}\mathbf{F}_2, \quad 
    \mathbf{F}_2 = \sum_{i=1}^N \frac{k_c}{N} (\mathbf{p}_{2,i}^* - \mathbf{p}_2),
    \label{eq:kinematic_novice}
\end{equation}
Here, $N = 10$ spots form a blurry point cloud. Each spot moves with a random velocity sampled from a Gaussian distribution (zero mean, $0.3\,$m/s standard deviation). This standard deviation matches the maximum target velocity used in \cite{TakagiBurdet2017}. We convert this visual disturbance into a haptic disturbance by assigning each blurred target point an independent virtual stiffness ($k_c/N$). The resulting variability is characterised by the Standard Error of Mean, corresponding to a normal distribution $\mathcal{N}(0,0.095^2)$.
      
\subsection{Experimental Protocol}
\noindent Experiments initially began in the Unconnected Mode (UM), which disabled all interactions between the robots. This baseline condition assessed individual motor performance. It also quantified the initial skill differences, as manipulated by the distribution method. Following the Unconnected Mode, the Connected Mode (CM) established the interactions through a virtual stiffness. In this mode, each condition applied identical stiffness along both axes ($k_x = k_y$), set to $0.5\KXY$, $\KXY$, $2\KXY$, or $4\KXY$. Time delay was simultaneously varied across $0$, $0.5\DXY$, $\DXY$, and $2\DXY$. The base parameters $\KXY$ and $\DXY$ are defined in \eqref{eq:mutiple}. Conditions are labelled as either UM or CM--$\mathrm{\alpha}$K--$\mathrm{\beta}$D. The $\alpha$ and $\beta$ multipliers denote the respective stiffness and delay levels. In Connected Mode, the stiffness levels correspond to $18$, $36$, $71$, or $142$\,N/m, and the delay levels correspond to $0$, $84$, $167$, or $334$\,ms. Each specific condition was repeated for twenty trials ($n = 20$) to ensure statistical reliability.

\subsection{Data Analysis}
\noindent Qualitatively, the raw data shown in Fig.~\ref{fig:raw_plot} illustrate the repeated HEBI trajectories mediated by H-MANs. For a quantitative assessment of the impact of haptic delay, we evaluated performance using tracking error (TE). TE is a standard metric to evaluate human motor performance \cite{DuCampolo2024,IvanovaBurdet2021,KagerCampolo2019}, enabling the quantification of performance into comparable data. Because the datasets are unpaired, the non-parametric Kruskal-Wallis test was applied to assess significant differences across multiple conditions. We utilised the non-parametric unpaired Wilcoxon rank sum test for pairwise comparisons.

\subsubsection*{Tracking Error (TE)}
Tracking Error is defined as the mean Euclidean distance between the actual trajectory and the target trajectory, and it has been used as a measurement of tracking accuracy \cite{DuCampolo2024, KagerCampolo2019}. In this study, the target is the preprogrammed position, as given by:
\begin{equation}
    \operatorname{TE_i} = \frac{1}{N}\sum_{n=1}^N \left\| \mathbf{p}^*(t_n) - \mathbf{p}_i(t_n) \right\|_2, \quad i = 1,2,
\end{equation}
where $n$ denotes the sample index, $N$ is the total number of samples, and $t_n$ is the $n$th sampled time point. $\mathrm{TE_1}$ and $\mathrm{TE_2}$ represent the tracking errors for HEBI\textsubscript{1} and HEBI\textsubscript{2}, respectively.

\section{Experimental Results}
\begin{figure}[!t]
    \centering
    \includegraphics[width=\columnwidth]{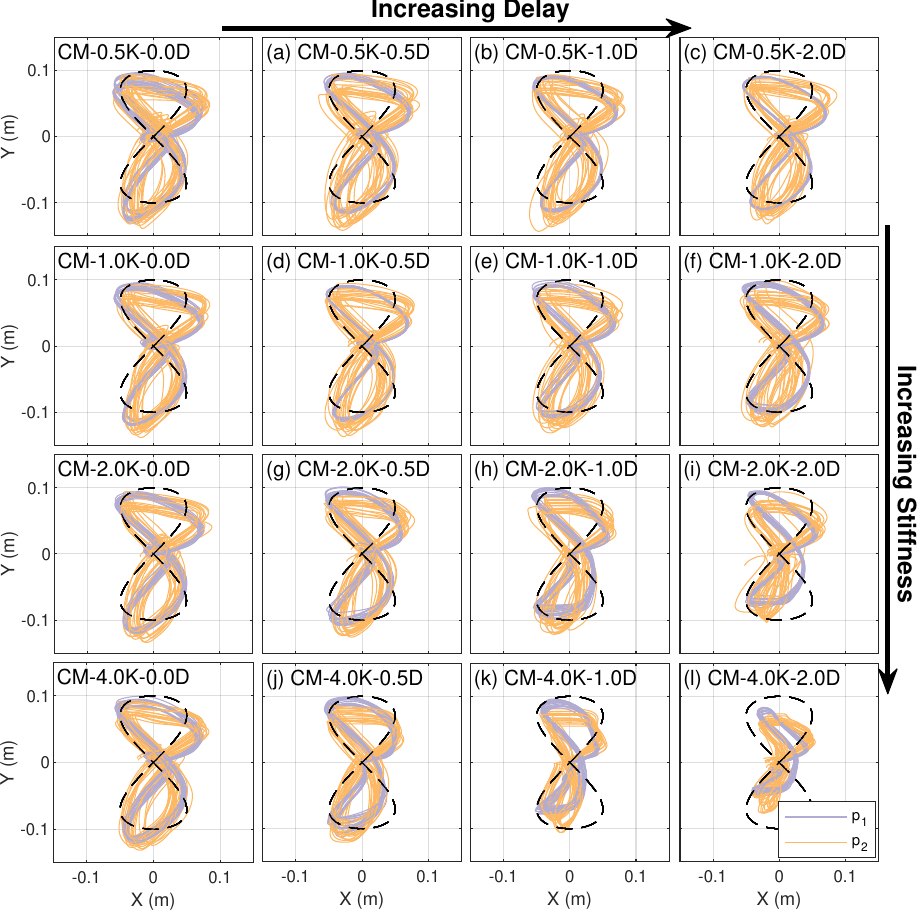}
    \caption{Position trajectories ($\mathbf{p}_1$ and $\mathbf{p}_2$ in m) of two H-MANs (H-MAN\textsubscript{1} and H-MAN\textsubscript{2}). The trajectory $\mathbf{p}_1$ corresponds to H-MAN\textsubscript{1} connected with the expert HEBI\textsubscript{1}, while $\mathbf{p}_2$ corresponds to H-MAN\textsubscript{2} connected with the novice HEBI\textsubscript{2}. CM--$\mathrm{\alpha}$K--$\mathrm{\beta}$D denotes the Connected Mode with stiffness set to $\alpha\KXY$ and delay set to $\beta\DXY$, respectively. Twenty trials ($n=20$) were repeated under each condition.}
    \label{fig:raw_plot}
\end{figure}
\begin{figure}[!t]
    \centering
    \includegraphics[width=\columnwidth]{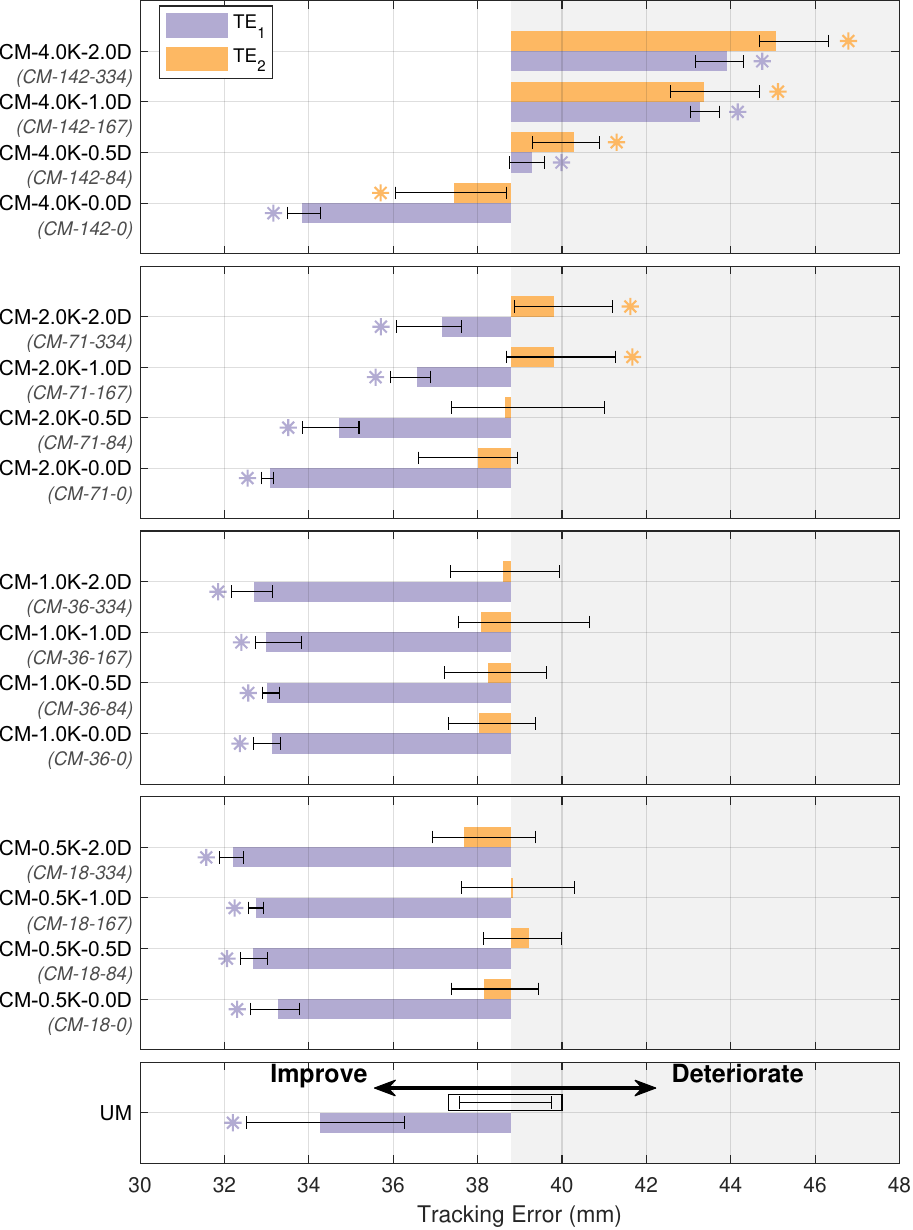}
    \caption{Tracking error (in mm for readability) under different experimental conditions: Unconnected Mode (UM) and Connected Mode (CM) with varying interaction stiffness and haptic delay levels. For each stiffness condition ($0.5\KXY$, $\KXY$, $2\KXY$, and $4\KXY$, corresponding to $18$, $36$, $71$, and $142$\,N/m), delays were systematically increased from $0$, $0.5\DXY$, $\DXY$, to $2\DXY$ ($0$, $84$, $167$, and $334$\,ms) to examine the effects of stiffness and delay on the tracking error of the expert HEBI\textsubscript{1} and novice HEBI\textsubscript{2} (TE\textsubscript{1} and TE\textsubscript{2}). TE\textsubscript{2} under UM ($38.75 \pm 1.69\,$mm) serves as the \fbox{baseline} for statistical comparisons evaluating performance improvement or deterioration. The asterisk (*) indicates $p < 0.05$ versus TE\textsubscript{2} under UM.}
    \label{fig:experiment_analysis}
\end{figure}
\noindent The trajectories of H-MAN\textsubscript{1} and H-MAN\textsubscript{2}, manipulated by HEBI\textsubscript{1} and HEBI\textsubscript{2} respectively, under different conditions are illustrated in Fig.~\ref{fig:raw_plot}. Qualitatively, at a lower stiffness of $0.5\KXY$, tracking performance shows no clear trend across varying haptic delays. However, at a high stiffness of $4\KXY$, the trajectories distort noticeably under the haptic delay of $2\DXY$. This implies that higher interactive stiffness increases system vulnerability to delay.

Quantitative analysis supports thesis visual observations (see Fig.~\ref{fig:experiment_analysis} and Table \ref{tab:experiment_results}). Pairwise comparisons evaluated via the non-parametric Wilcoxon rank-sum test reveal that in the Unconnected Mode, HEBI\textsubscript{1} outperforms HEBI\textsubscript{2} significantly ($p<0.001$, pairwise comparison of TE\textsubscript{1} and TE\textsubscript{2} under UM). This demonstrates that the haptic disturbance substantially degraded the initial skill of HEBI\textsubscript{2}. Arrows in Table \ref{tab:experiment_results} denote significant deviations from the TE\textsubscript{2} Unconnected Mode baseline ($p<0.05$ for pairwise comparisons with TE\textsubscript{2} under UM). Interacting with the expert via a high stiffness of 142\,N/m improves HEBI\textsubscript{2} performance over isolated operation (TE\textsubscript{2}: $p<0.05$ for CM-142-0 versus UM). However, introducing delays under this high stiffness significantly deteriorates performance (TE\textsubscript{2}: $p<0.05$ for all CM-142 delay conditions versus UM).

At lower stiffness levels of 18 and 36\,N/m, HEBI\textsubscript{2} showed no significant performance improvement. Furthermore, introducing delays at these levels had no measurable impact ($p>0.05$ for multiple comparisons across UM and corresponding CM conditions). In contrast, at higher stiffness levels of 71 or 142\,N/m, haptic delays negatively affected performance (TE\textsubscript{2}: $p<0.001$ for multiple comparisons among CM-71 or CM-142 conditions). Notably, a stiffness of 71\,N/m without delay did not significantly improve HEBI\textsubscript{2}'s performance (TE\textsubscript{2}: $p>0.05$ for CM-71-0 versus UM). Yet, adding delays of 167 or 334\,ms to this stiffness severely degraded performance (TE\textsubscript{2}: $p<0.001$ for CM-71-167 or CM-71-334 versus UM). There was no significant difference between the effects of 167 and 334\,ms delays at 71\,N/m (TE\textsubscript{2}: $p>0.05$ for CM-71-167 versus CM-71-334). However, at the maximum stiffness of 142\,N/m, the performance progressively worsened as delays increased from 84, 167, to 334\,ms (TE\textsubscript{2}: $p<0.01$ for CM-142-0 versus CM-142-84, CM-142-84 versus CM-142-167, and CM-142-167 versus CM-142-334). These results indicate that haptic delay hinders dyadic interaction by degrading the performance of novice participants, potentially causing them to perform even worse than in solo trials. While lower stiffness reduces the dyadic system's vulnerability to haptic delay, it fails to provide cooperative performance benefits.
\begin{table}[!t]
\renewcommand{\arraystretch}{1.2}
\caption{Tracking Error (TE) \\
under varying stiffness and delay conditions.}
\label{tab:experiment_results}
\begin{center}
\begin{tabular}{m{8em} m{8em} m{8em}}
\hline
TASK & TE\textsubscript{1} (mm) & TE\textsubscript{2} (mm)\\
\hline
\vspace{0.3em}
UM           & $34.83 \pm 2.72 \downarrow$ & \fbox{$38.75 \pm 1.69$} \\
CM-18-0      & $33.24 \pm 0.69 \downarrow$ & $38.45 \pm 1.36$ \\
CM-18-84     & $32.63 \pm 0.56 \downarrow$ & $39.36 \pm 1.80$ \\
CM-18-167    & $32.72 \pm 0.37 \downarrow$ & $39.30 \pm 2.33$ \\
CM-18-334    & $32.11 \pm 0.71 \downarrow$ & $37.90 \pm 2.12$ \\
CM-36-0      & $33.01 \pm 0.38 \downarrow$ & $38.25 \pm 1.67$ \\
CM-36-84     & $32.99 \pm 0.59 \downarrow$ & $38.23 \pm 1.62$ \\
CM-36-167    & $33.19 \pm 0.61 \downarrow$ & $39.03 \pm 2.11$ \\
CM-36-334    & $32.69 \pm 0.62 \downarrow$ & $39.10 \pm 2.83$ \\
CM-71-0      & $33.06 \pm 0.33 \downarrow$ & $38.17 \pm 1.72$ \\
CM-71-84     & $34.45 \pm 1.16 \downarrow$ & $38.93 \pm 2.12$ \\
CM-71-167    & $36.53 \pm 0.69 \downarrow$ & $40.04 \pm 2.08 \uparrow$ \\
CM-71-334    & $36.84 \pm 1.10 \downarrow$ & $40.27 \pm 2.19 \uparrow$ \\
CM-142-0     & $33.88 \pm 0.73 \downarrow$ & $37.75 \pm 2.15 \downarrow$ \\
CM-142-84    & $39.19 \pm 0.66 \uparrow$   & $40.11 \pm 0.96 \uparrow$ \\
CM-142-167   & $43.28 \pm 0.80 \uparrow$   & $43.65 \pm 1.36 \uparrow$ \\
CM-142-334   & $43.84 \pm 0.98 \uparrow$   & $45.45 \pm 1.11 \uparrow$ \\
\hline
\end{tabular}
\end{center}
\vspace{0.3em}
\footnotesize{
Data are shown as mean $\pm$ standard deviation in units of (mm).\\
\fbox{Baseline} corresponds to TE\textsubscript{2} under the Unconnected Mode and serves as the reference for significance comparison.\\ $\uparrow$ or $\downarrow$ indicates the corresponding metric measured under the corresponding condition is significantly greater or less than the baseline condition ($p < 0.05$ for pairwise comparison with TE\textsubscript{2} in UM).}
\end{table}

\section{Discussion}
\noindent This study established analytical stability criteria for delayed robot-mediated dyadic interactions. We systematically examined how this theoretical stability correlates with physical motor performance. Frequency-domain analyses reveal that the dyadic system achieves delay-independent stability when interactive stiffness remains below a critical threshold corresponding to system dynamics. Additionally, we identified a maximum tolerable delay boundary ($\delta_m$) corresponding to system dynamics and controllers. Any delay below this specific threshold guarantees system stability. Consequently, sub-critical stiffness ($\leq k_m$) ensures absolute robustness against unpredictable network delays. In contrast, super-critical stiffness ($> k_m$) demands strict delay regulation to stay within the $\delta_m$ boundary. These stability thresholds depend entirely on the inherent robot dynamics, namely inertia and damping. Damping dominants this relationship due to its quadratic effect on the stability margin. In practical applications with commercial rehabilitation platforms (H-MAN), this suggests a trade-off: low-stiffness connections tolerate substantial delays, whereas high-stiffness connections require strict delay regulation.

We extended these stability criteria to multi-DOF configurations to enable applicability across diverse robotic platforms. We subsequently modelled a remote interaction between operators with different skill levels (e.g., a therapist and a patient). This was achieved using two human-like HEBI-based robots executing preprogrammed, human-inspired trajectories. Experimental results verified the theoretical predictions. Under sub-critical stiffness ($0.5k_m$ or $k_m$), haptic delays did not significantly affect novice-like robot's performance. This is consistent with the analytical delay-independent stability prediction. Conversely, under super-critical stiffness ($2k_m$ or $4k_m$), the novice-like robot is hindered by elevated delays ($\delta_m$ or $2\delta_m$). 
This performance degradation consistently occurs when interaction parameters cross into mathematically unstable regions, suggesting that theoretical system stability is a fundamental prerequisite for effective physical cooperation. We define ``hinder'' as cooperative performance falling below isolated solo performance \cite{IvanovaBurdet2021}. This highlights that combining high stiffness with haptic delay renders dyadic interaction physically counterproductive. These empirical findings corroborate previous behavioural studies \cite{DuCampolo2024, IvanovaBurdet2021}, and reinforce that delay-induced instability dictates performance deterioration in physical cooperation.

Notably, lower experimental stiffness levels ($0.5k_m$, $1.0k_m$, and $2k_m$) provided no measurable cooperative benefits, even without delay. The critical stiffness $k_m$ is proportional to the squared damping coefficient ($k_m \propto b^2$). Because the H-MAN exhibits low intrinsic damping, the absolute stiffness at $2k_m$ ($71$\,N/m) remained too compliant to transfer interactive assistance. This can also be inferred from the maximum stiffness condition ($4k_m$ or $142$\,N/m). Only at this elevated stiffness did the novice-like robot exhibit significant performance improvements. Therefore, mere system stability does not guarantee a clinically beneficial interaction. Effective cooperation often requires a highly stiff connection to transmit guiding forces, as demonstrated in this study using rehabilitation robots. However, this high stiffness drastically increases the system’s vulnerability to haptic delay. This fundamental trade-off suggests the implementation of stability-aware buffer mechanism. Active delay compensation is strictly necessary to maintain safety while rendering the stiff environments required for motor training or resistive rehabilitation 

The limitations of this study include the use of human-like robots as substitutes for human participants. Future experiments involving real humans are needed to clarify the correlation between system stability and motor performance in human-human interactions. In this study, H-MAN was modelled as a linear damping system with constant damping throughout the movement; adopting a complex, non-linear model could enhance the accuracy of instability predictions. Haptic delay was assumed to be constant, whereas network delays can vary over time. Future work should address dyadic interactions under time-varying haptic delays to better match real-world conditions. Overall, the derived stability criteria can guide the design of stability-aware buffer systems for delay compensation. For broader applications, stiffness levels should be carefully chosen relative to the robot’s dynamic properties to ensure effective and stable remote dyadic interactions.

\section{Conclusion}
\noindent This study developed a dynamic model to represent remote, robot-mediated dyadic interactions over delayed networks. We applied the exact zero-crossing methodology to rigorously define the boundary between stable and unstable interaction regions. This work establishes explicit, analytical stability criteria for delayed haptic coupling. These criteria derive directly from identified hardware parameters, providing a rigorous mathematical foundation for designing active delay-buffer systems.

Empirical validation using human-like robots demonstrate that system instability directly dictates performance deterioration in dyadic interactions. Under unstable delayed conditions, the interaction can even become counterproductive, leading to poorer outcomes than those observed during solo performance. These findings underscore the necessity of constraining interactions within the stability region to ensure a safe and effective cooperation. Overall, this framework provides essential insights for deploying remote healthcare robotics and synthesising exact delay compensation mechanisms.

\bibliographystyle{IEEEtran}
\bibliography{reference}
\vspace{1em}
{\appendix[]
}
\subsection{Analytical Proof for Direct Coupling Stability} \label{appendix: coupling}
\begin{proof}
The leading coefficient $C_6 = m_1^2 m_2^2$ is strictly positive, ensuring $\lim_{\omega \to +\infty} \mathcal{F}(\omega) = +\infty$. At $\omega = 0$, the polynomial evaluates to $\mathcal{F}(0) = C_0$. According to Descartes' Rule of Signs, since $C_6 > 0$, the necessary and sufficient condition to rigorously preclude the existence of any positive real roots is that all subsequent polynomial coefficients must remain strictly positive ($C_4 > 0 \;\land\; C_2 > 0 \;\land\; C_0 > 0$). By enforcing positivity across these three constraints independently, we can derive the exact boundaries for the stiffness $k$:

1) From $C_0 > 0$, we establish the base critical stiffness:
\begin{equation}
    k < k_{c0} = \frac{b_1^2 + b_2^2}{2(m_1 + m_2)}.
\end{equation}

2) From $C_4 > 0$, we extract the mass-weighted boundary:
\begin{equation}
    k < k_{c4} = \frac{b_1^2 m_2^2 + b_2^2 m_1^2}{2 m_1 m_2 (m_1 + m_2)}.
\end{equation}

3) From $C_2 > 0$, expanding the terms yields a quadratic equation with respect to $k$, with a reduced discriminant defined as:
\begin{equation}
    \Lambda = \left[ (b_1^2 - b_2^2)(m_2^2 b_1^2 - m_1^2 b_2^2) - 2m_1 m_2 b_1^2 b_2^2 \right].
\end{equation}
When $\Lambda \ge 0$, ensuring $C_2 > 0$ imposes an additional critical resonance boundary:
\begin{equation}
    k < k_{c2} = \frac{m_1 b_2^2 + m_2 b_1^2 - \sqrt{\Lambda}}{m_1^2 + 4m_1 m_2 + m_2^2}.
\end{equation}
Conversely, if $\Lambda < 0$, the algebraic root for $k_{c2}$ becomes complex, indicating that $C_2$ remains strictly positive for all real values of $k$. Therefore, to prevent instability, the interactive stiffness $k$ must satisfy the most restrictive of these constraints:
\begin{equation}
    k \le k_{m} \equiv \mathcal{S}(m_1, b_1, m_2, b_2) = \min \big( \{k_{c0}, k_{c2}, k_{c4}\} \cap \mathbb{R} \big).
\end{equation}
\end{proof}

\subsection{Virtual Proxy Characteristic Polynomial Expansion}
\label{appendix: proxy}
\noindent When the virtual proxy dynamics ($m_p$, $b_p$) are incorporated into the delay channel alongside independent forward and return coupling stiffnesses ($k_1$, $k_2$), the characteristic frequency equation expands into a highly complex 10th-order polynomial, $\mathcal{F}_{p}(\omega)$, and a correspondingly extended complex delay parameter, $z_{p}(\omega)$. The explicit algebraic expansions governing this proxy-mediated dyadic coupling are formulated in \eqref{eq:proxy_poly} and \eqref{eq:proxy_z}.
\begin{figure*}[!t]
\small
\hrulefill
\begin{equation}
\mathcal{F}_{p}(\omega) = C_{10}\omega^{10} + C_8\omega^8 + C_6\omega^6 + C_4\omega^4 + C_2\omega^2 + C_0 = 0,
\label{eq:proxy_poly}
\end{equation}
where the coefficients $C_{2i}$ are analytically derived as:
\begin{align*}
    C_{10} &= m_1^2 m_2^2 m_p^2, \\
    C_8 &= - 2k_2 m_1^2 m_2 m_p (m_2 + m_p) - 2k_1 m_1 m_2^2 m_p (m_1 + m_p) + b_p^2 m_1^2 m_2^2 + b_2^2 m_1^2 m_p^2 + b_1^2 m_2^2 m_p^2, \\
    C_6 &= 4k_1 k_2 m_1 m_2 m_p (m_1 + m_2 + m_p) + 4k_2^2 m_1^2 m_2 m_p + 2k_1^2 m_1 m_2^2 m_p + 2k_1 k_2 m_1^2 m_2^2 - 2b_p^2 k_2 m_1^2 m_2 - 2b_2^2 k_2 m_1^2 m_p \\
           &\quad - 2b_1^2 k_2 m_2 m_p (m_2 + m_p) - 2b_p^2 k_1 m_1 m_2^2 - 2b_2^2 k_1 m_1 m_p (m_1 + m_p) - 2b_1^2 k_1 m_2^2 m_p + k_2^2 m_1^2 (m_p^2 + m_2^2) \\
           &\quad + k_1^2 m_2^2 (m_p^2 + m_1^2) + b_2^2 b_p^2 m_1^2 + b_1^2 b_p^2 m_2^2 + b_1^2 b_2^2 m_p^2, \\
    C_4 &= - 8k_1 k_2^2 m_1 m_2 m_p - 4k_1^2 k_2 m_1 m_2 m_p + 4b_p^2 k_1 k_2 m_1 m_2 + 4b_2^2 k_1 k_2 m_1 m_p + 4b_1^2 k_1 k_2 m_2 m_p - 2k_1^2 k_2 m_2 m_p (m_2 + m_p) \\
           &\quad - 2k_1 k_2^2 m_1 m_p (m_1 + m_p) - 4k_1 k_2^2 m_1^2 m_2 - 2k_1^2 k_2 m_1 m_2 (m_1 + m_2) - 2k_1 k_2^2 m_1 m_2^2 + 4b_1^2 k_2^2 m_2 m_p \\
           &\quad + 2b_2^2 k_1^2 m_1 m_p + 2b_2^2 k_1 k_2 m_1^2 + 2b_1^2 k_1 k_2 m_2^2 - 2b_1^2 b_p^2 k_2 m_2 - 2b_1^2 b_2^2 k_2 m_p - 2b_2^2 b_p^2 k_1 m_1 - 2b_1^2 b_2^2 k_1 m_p \\
           &\quad + 2b_1 b_p k_1^2 m_2^2 - 2k_2^3 m_1^2 (m_p + m_2) + b_p^2 (k_2^2 m_1^2 + k_1^2 m_2^2) + b_2^2 (k_1^2 m_p^2 + k_2^2 m_1^2 + k_1^2 m_1^2) \\
           &\quad + b_1^2 (k_2^2 m_p^2 + k_2^2 m_2^2 + k_1^2 m_2^2) + b_1^2 b_2^2 b_p^2, \\
    C_2 &= - 4b_1 b_p k_1^2 k_2 m_2 + 4k_1 k_2^3 m_1 (m_p + m_2) + 4k_1^2 k_2^2 m_1 m_2 + 2k_1^2 k_2^2 m_p (2m_2 + m_1) - 2b_p^2 k_1 k_2 (k_1 m_2 + k_2 m_1) \\
           &\quad - 2b_2^2 k_1 k_2 (k_1 m_p + k_1 m_1 + k_2 m_1) - 2b_1^2 k_1 k_2^2 (m_p + 2m_2) - 2b_1^2 k_1^2 k_2 m_2 + 2b_1^2 b_2^2 k_1 k_2 + 2b_1 b_2^2 b_p k_1^2 \\
           &\quad + 2k_1 k_2^3 m_1^2 - 2b_1^2 k_2^3 (m_p + m_2) + k_1^2 k_2^2 (m_p^2 + m_2^2 + m_1^2) + b_2^2 b_p^2 k_1^2 + b_1^2 b_p^2 k_2^2 + b_1^2 b_2^2 (k_2^2 + k_1^2) + 2b_1 b_p k_1^2 k_2^2, \\
    C_0 &= - 2k_1^2 k_2^3 (m_p + m_2 + m_1) + 2b_1^2 k_1 k_2^3 + k_1^2 k_2^2 (b_p^2 + b_2^2 + b_1^2).
\end{align*}
Once a valid crossing frequency $\omega$ is identified, the corresponding exact delay parameter $z_{p}(\omega)$ is evaluated at $s = j\omega$ to isolate the complex delay boundary. The formulation can be distributed and simplified as:
\begin{equation}
    z_{p}(\omega) = \frac{1}{k_2^2} \left[ (-m_p\omega^2 + jb_p\omega + k_1 + k_2) - \frac{k_1^2}{-m_1\omega^2 + jb_1\omega + k_1} \right] (-m_2\omega^2 + jb_2\omega + k_2).
    \label{eq:proxy_z}
\end{equation}
\hrulefill
\vspace*{4pt}
\end{figure*}

\subsection{Notation and Numerical Values}
\begin{table}[H]
    \caption{Notations and Corresponding Numerical Values\label{tab:notation}}
    \begin{center}
    \begin{tabular}{c c c m{16em}}
    \hline
    Notation &  Value &  Unit & Note\\
    \hline
    $\mathrm{M}$ & 0.8334 & $\mathrm{kg}$ & Base mass (inertia)\\
    $\mathrm{B}$ & 7.7257 & $\mathrm{Ns/m}$ & Base damping coefficient\\
    $\mathrm{K}$ & 36 & $\mathrm{N/m}$ & Base stiffness for 1 DOF Simulation\\
    $\mathrm{\Delta}$ & 169 & $\mathrm{ms}$ & Base delay for 1 DOF Simulation\\
    $\mathrm{M_1^x}$ & 0.8334 & $\mathrm{kg}$ & Estimated mass of H-MAN\textsubscript{1} x-axis\\    $\mathrm{M_1^y}$ & 1.0649 & $\mathrm{kg}$ & Estimated mass of H-MAN\textsubscript{1} y axis\\    $\mathrm{M_2^x}$ & 0.7776 & $\mathrm{kg}$ & Estimated mass of H-MAN\textsubscript{2} x-axis\\    $\mathrm{M_2^y}$ & 1.3407 & $\mathrm{kg}$ & Estimated mass of H-MAN\textsubscript{2} y axis\\
    $\mathrm{B_1^x}$ & 7.7257 & $\mathrm{Ns/m}$ & Estimated damper of H-MAN\textsubscript{1} x-axis\\
    $\mathrm{B_1^y}$ & 10.1168 & $\mathrm{Ns/m}$ & Estimated damper of H-MAN\textsubscript{1} y-axis\\
    $\mathrm{B_2^x}$ & 7.4208 & $\mathrm{Ns/m}$ & Estimated damper of H-MAN\textsubscript{2} x-axis\\
    $\mathrm{B_2^y}$ & 9.3496 & $\mathrm{Ns/m}$ & Estimated damper of H-MAN\textsubscript{2} y-axis\\
    $\KXY$ & 36 & $\mathrm{N/m}$ & Reference stiffness\\
    $\mathrm{\DXY}$ & 165 & $\mathrm{ms}$ & Reference delay\\
    \hline
    \end{tabular}     
    \end{center}
\end{table}

\subsection{Dynamic Identification}
\begin{figure}[H]
    \centering
    \includegraphics[width=0.9\columnwidth]{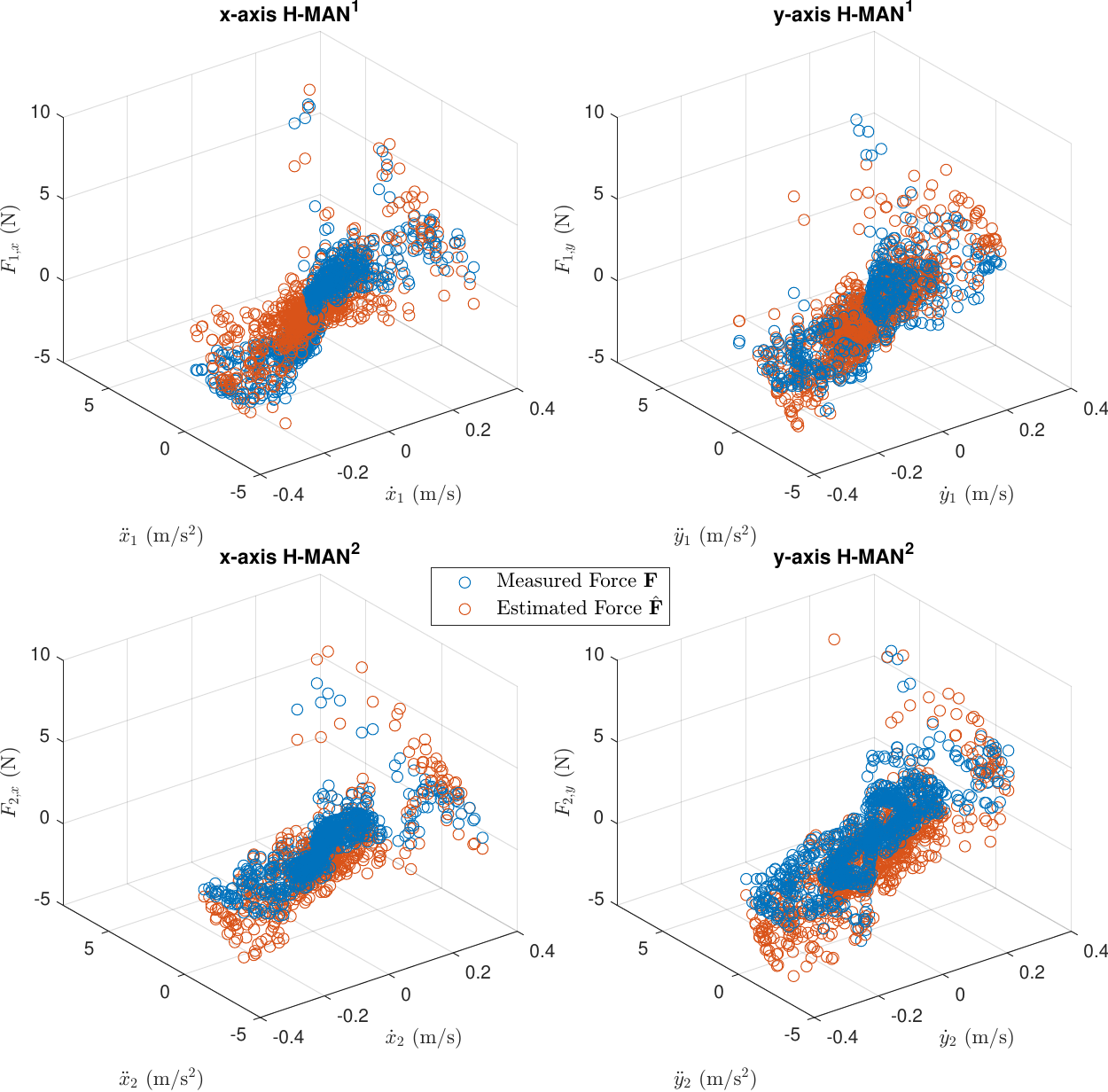}
    \caption{Estimated and measured forces, and their regression with velocity and acceleration. The estimated forces are obtained from $\hat{\mathbf{F}} = \mathbf{X}\hat{\boldsymbol{\beta}}$. The estimated force plane $\hat{\mathbf{F}}$ qualitatively aligns with the measured data plane $\mathbf{F}$, indicating that the identified dynamic parameters $\hat{\boldsymbol{\beta}}$ are reasonable.}
    \label{fig:leastSquare}
\end{figure}

\noindent System identification requires persistent signal excitation, classically achieved via Fourier series \cite{SweversVanBrussel1997}. Base parameters are subsequently extracted using optimisation techniques, such as weighted least squares estimation \cite{WuYou2010}. We designed the excitation trajectory based on \cite{SweversVanBrussel1997}. It is formulated as:
\[
\begin{split}
    \dot{x}_{\operatorname{cmd}}(t) &= \sum_{i=1}^N A_icos(i\omega t) + B_isin(i\omega t),\\
    \dot{y}_{\operatorname{cmd}}(t) &= \sum_{i=1}^N A_icos(i\omega t) + B_isin(i\omega t),
\end{split}
\]
where $N=5$, $\mathbf{A} = [0.01,\,0.02,\,0.05,\,0.1,\,0.15]^{\top}$, and $\mathbf{B} = [0.01,\,0.02,\,0.05,\,0.1,\,0.15]^{\top}$.
We estimate parameters by collecting $n$ discrete data points to construct the matrix equation:
\[
    \mathbf{F} = \mathbf{X}\boldsymbol{\beta},\quad \begin{bmatrix}
        F_1\\
        F_2\\
        \vdots\\
        F_n
    \end{bmatrix} = \begin{bmatrix}
        \ddot{x}_1 & \dot{x}_1\\
        \ddot{x}_2 & \dot{x}_2\\
        \vdots & \vdots\\
        \ddot{x}_n & \dot{x}_n\\
    \end{bmatrix}\begin{bmatrix}
        m\\
        b
    \end{bmatrix},
\]
where we can solve for $m$ and $b$ through least squares,
\[
    \boldsymbol{\beta} = (\mathbf{X}^{\top}\mathbf{X})^{-1}\mathbf{X}^{\top}\mathbf{F}.
\]
The weighted least squares method can be expressed as:
\[
  \boldsymbol{\hat{\beta}}= \arg\min_{\boldsymbol{\beta}}(r^\top \mathbf{W} r),\quad r = \mathbf{F} - \mathbf{X}\boldsymbol{\beta},
\]
where $\mathbf{W}$ is a diagonal weight matrix with entries $w_{ii} = \frac{1}{r^2 + \epsilon}$. The scalar $\epsilon$ is the floating-point relative accuracy to avoid division by zero. Therefore, the 2-DOF linear dynamic parameters (for the $\operatorname{x}$ and $\operatorname{y}$ axes) of both H-MAN\textsubscript{1} and H-MAN\textsubscript{2} can be estimated. The final identified parameters are detailed in Table \ref{tab:notation}, and the resulting regression planes are visualised in Fig.~\ref{fig:leastSquare}.
\end{document}